# Gyroscope Calibration via Magnetometer

Yuanxin Wu, *Senior Member of IEEE,* and Ling Pei

*Abstract*—Magnetometers, gyroscopes and accelerometers are commonly used sensors in a variety of applications. The paper proposes a novel gyroscope calibration method in the homogeneous magnetic field by the help of magnetometer. It is shown that, with sufficient rotation excitation, the homogeneous magnetic field vector can be exploited to serve as a good reference for calibrating low-cost gyroscopes. The calibration parameters include the gyroscope scale factor, non-orthogonal coefficient and bias for three axes, as well as its misalignment to the magnetometer frame. Simulation and field test results demonstrate the method's effectiveness.

*Index Terms*—Magnetometer calibration, gyroscope calibration, accelerometer, misalignment

## I. INTRODUCTION

Magnetometers, gyroscopes and accelerometers have nowadays become commonly used sensors in a variety of applications including but not limited to pedestrian indoor navigation, flight stabilization, satellite attitude, augmented reality, human body motion tracking and medical instrument [2-4]. Magnetometers measure the ambient magnetic field; gyroscopes and accelerometers respectively sense the angular velocity and non-gravitational acceleration of the rigidly-attached platform. The latter two are collectively known as inertial sensors. With the advancing MEMS technology, these three-axis sensors have been commonly integrated into an all-in-one compact and low cost sensor module or chip [5]. A conveniently-implemented quality calibration is desirable to guarantee their proper performance in each application. A reference input is a fundamental requirement of any sensor calibration [6], for instance, using the local gravity for accelerometers [7] or using the local geomagnetic field for magnetometers [1, 8, 9]. These two reference inputs are physical quantities that naturally exist on the Earth. It is, however, quite cumbersome to find a reference sensor input for gyroscopes, as the Earth rotation rate is too small in magnitude to be used for consumer sensors. Existing gyroscope calibration methods mostly rely on external apparatus to provide reference inputs, such as a turn rate table [10] or a bicycle wheel [11].

This paper proposes a novel method for gyroscope calibration, aided by the magnetometer in the same module or chip. The calibration process is supposed to be done in a homogeneous magnetic field. That is to say, the magnetic field vector does not need to be known but should be fixed. It consists of two steps that are sequentially performed on the same data: 1) the magnetometer interior calibration is performed using the fact the norm of the magnetic field is invariant; 2) with the aid of the calibrated magnetometer, the gyroscope parameters are calibrated, as well as its orientation misalignment to the magnetometer. The gyroscope parameters include bias, scale factor and non-orthogonal coefficient for three axes. Sufficient rotation excitation (covering rotation about two or more axes) is needed to get a good calibration result, but the specific form of rotation excitation is not restricted. Better calibration quality would be obtained with more sufficient rotation excitation.

The paper is organized as follows. Section II describes the sensor models of magnetometers and gyroscopes. Section III formulates the gyroscope calibration problem as a state-space model and examines its observability property. Section IV reports simulation and field test results and the conclusions are drawn in Section V.

## II. MAGNETOMETER AND GYROSCOPE

### A. Magnetometer

Taking the time-invariant magnetic disturbance and sensor imperfection into account, the three-axis magnetometer measurement can be collectively modelled by [1, 12, 13]

$$\mathbf{y}_m = \mathbf{R}^{-1}\mathbf{Q}^T\mathbf{C}_e^{m^*}\mathbf{m}^e + \mathbf{h} \triangleq \mathbf{R}^{-1}\mathbf{m}^m + \mathbf{h} \quad (1)$$

where $\mathbf{m}^e$ is the typically unknown local magnetic vector in the Earth frame, the superscript $m^*$ means the magnetometer sensor frame defined by the physical axes and $\mathbf{C}_e^{m^*}$ is the orientation matrix of $m^*$-frame relative to e-frame. For a homogenous magnetic field at the

This work was supported in part by National Natural Science Foundation of China (61174002, 61422311) and Hunan Provincial Natural Science Foundation of China (2015JJ1021).

Authors' address: Shanghai Key Laboratory of Navigation and Location-based Services, School of Electronic Information and Electrical Engineering, Shanghai Jiao Tong University, Shanghai, China, 200240, E-mail: (yuanx_wu@hotmail.com).

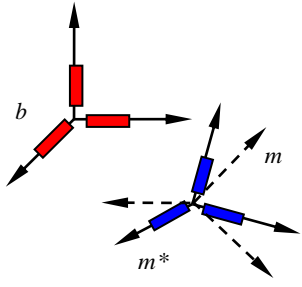

Figure 1. Gyroscope triad, magnetometer triad and their sensor frames. Filled blocks mean separate sensors. Solid frames (b-frame and $m^*$-frame) mean sensor frames defined by physical sensitivity axes and the dashed frame (m-frame) is the equivalent magnetometer frame induced by soft-iron effect. [1]

calibration site, $\mathbf{m}^e$ is constant and assumed to have unity norm without loss of generality. The superscript $m$ means the equivalent magnetometer frame, into which the magnetometer sensor frame $m^*$ is distorted by the soft-iron magnetic material onboard (see Fig. 1). The attitude matrix $\mathbf{Q}$ is the attitude discrepancy between these two frames and $\mathbf{R}$ is an upper triangular matrix that encodes the distorted scale factor matrix and the non-orthogonality matrix. Equation (1) can be written as the popular intrinsic magnetometer calibration model

$$\mathbf{m}^m = \mathbf{R}(\mathbf{y}_m - \mathbf{h}), \text{ with } \|\mathbf{m}^m\| = \|\mathbf{Q}^T \mathbf{C}_e^{m^*} \mathbf{m}^e\| = 1 \quad (2)$$

The estimate of the intrinsic parameters, $\hat{\mathbf{R}}$ and $\hat{\mathbf{h}}$, can be obtained by the iterative Newton method with a good initial estimate, cf. e.g. [14]. The calibrated magnetometer measurements are given as

$$\hat{\mathbf{m}}^m = \hat{\mathbf{R}}(\mathbf{y}_m - \hat{\mathbf{h}}) = \mathbf{m}^m + \mathbf{n}_m \quad (3)$$

where $\mathbf{n}_m$ denotes the magnetometer calibration residual that can be roughly modelled by Gaussian noise.

*B. Gyroscope and Orientation*

The gyroscope triad can measure the body angular velocity relative to the inertial space, expressed in the gyroscope frame

$$\mathbf{y}_g = \mathbf{K}_g^{-1}(\boldsymbol{\omega}_{ib}^b - \boldsymbol{\varepsilon}_b) + \mathbf{n}_g \quad (4)$$

where the superscript $b$ denotes the gyroscope frame defined by the physical sensor axes, $\boldsymbol{\omega}_{ib}^b$ is the true angular velocity, the upper triangular $\mathbf{K}_g^{-1}$ encodes the scale factor and the attitude misalignment between the gyroscope sensors, $\boldsymbol{\varepsilon}_b$ the random constant gyroscope bias and $\mathbf{n}_g$ the gyroscope measurement noise. The calibrated gyroscope measurement can be expressed as

$$\boldsymbol{\omega}_{ib}^b = \mathbf{K}_g(\mathbf{y}_g - \mathbf{n}_g) + \boldsymbol{\varepsilon}_b \triangleq \mathbf{K}_g \mathbf{y}_g + \boldsymbol{\varepsilon}_b + \mathbf{n}_b \quad (5)$$

The gyroscope frame's orientation with respect to the inertial frame can be computed by integrating the calibrated gyroscope measurements [6]

$$\dot{\mathbf{C}}_b^i = \mathbf{C}_b^i \boldsymbol{\omega}_{ib}^b \times = \mathbf{C}_b^i (\mathbf{K}_g \mathbf{y}_g + \boldsymbol{\varepsilon}_b + \mathbf{n}_b) \times \quad (6)$$

The skew symmetric matrix $(\cdot\times)$ is defined so that the cross product satisfies $\mathbf{x} \times \mathbf{y} = (\mathbf{x}\times)\mathbf{y}$ for arbitrary two vectors.

### III. MAGNETOMETER-AIDED GYROSCOPE CALIBRATION

*A. Problem Formulation*

Whenever the triads of magnetometer and gyroscope are strapped together, there will be an attitude misalignment between their respective frames. For instance, the calibrated magnetometer measurements can be re-expressed as

$$\hat{\mathbf{m}}^m = \mathbf{C}_i^m \mathbf{m}^i + \mathbf{n}_m = \mathbf{C}_b^m \mathbf{C}_i^b \mathbf{m}^i + \mathbf{n}_m \quad (7)$$

where $\mathbf{C}_b^m$ denotes the attitude misalignment between the magnetometer and gyroscope frames, which is constant once the sensors (and the magnetic material onboard) have been fixed onto the platform. Using (6), the change rate of the magnetometer frame's orientation with respect to the inertial frame is characterized by

$$\begin{aligned}
\dot{\mathbf{C}}_m^i &= \mathbf{C}_b^i \left[ (\mathbf{K}_g \mathbf{y}_g + \boldsymbol{\varepsilon}_b + \mathbf{n}_b) \times \right] \mathbf{C}_m^b \\
&= \mathbf{C}_m^i \left[ \mathbf{C}_b^m (\mathbf{K}_g \mathbf{y}_g + \boldsymbol{\varepsilon}_b + \mathbf{n}_b) \right] \times \\
&\triangleq \mathbf{C}_m^i (\mathbf{K} \mathbf{y}_g + \boldsymbol{\varepsilon} + \mathbf{n}) \times
\end{aligned} \quad (8)$$

where $\mathbf{K} \triangleq \mathbf{C}_b^m \mathbf{K}_g$, $\boldsymbol{\varepsilon} = \mathbf{C}_b^m \boldsymbol{\varepsilon}_b$ and $\mathbf{n} = \mathbf{C}_b^m \mathbf{n}_b$. Note that the misalignment $\mathbf{C}_b^m$ and the upper triangular $\mathbf{K}_g$ can be recovered by the orthogonal-triangular (QR) decomposition of the fully-populated matrix $\mathbf{K}$.

Then the gyroscope calibration problem can be formulated as a state-space model which takes (8) as the dynamic model and (7) as the observation model. Specifically,

$$\begin{cases}
\dot{\mathbf{C}}_m^i = \mathbf{C}_m^i (\mathbf{K} \mathbf{y}_g + \boldsymbol{\varepsilon} + \mathbf{n}) \times \\
\dot{\mathbf{K}} = 0 \\
\dot{\boldsymbol{\varepsilon}} = 0 \\
\dot{\mathbf{m}}^i \approx 0 \\
\hat{\mathbf{m}}^m = \mathbf{C}_i^m \mathbf{m}^i + \mathbf{n}_m
\end{cases} \quad (9)$$

The magnitude of the vector rate $\|\dot{\mathbf{m}}^i\| = \|\mathbf{C}_e^i (\boldsymbol{\omega}_{ie}^e \times) \mathbf{m}^e\|$.

As the magnitude of the Earth's rotation rate ($\Omega \approx 7.3 \times 10^{-5}$ rad/s) is much smaller than low-cost gyroscope errors, $\|\dot{\mathbf{m}}^i\| < \Omega$ and it is reasonable to roughly take $\dot{\mathbf{m}}^i$ as zero. The states to be estimated include the magnetometer orientation $\mathbf{C}_m^i$, the gyroscope parameters $\mathbf{K}$ and $\boldsymbol{\varepsilon}$, and the local magnetic vector $\mathbf{m}^i$. Then the gyroscope parameters can be retrieved by $[\mathbf{C}_b^m, \mathbf{K}_g] = qr(\mathbf{K})$ and $\boldsymbol{\varepsilon}_b = \mathbf{C}_m^b \boldsymbol{\varepsilon}$, where the operator $qr(\cdot)$ means the QR decomposition.

*B. Observability Property*

*Definition of State Observability* [15]: A system is said to be (globally) observable if for any unknown initial state $\mathbf{x}(0)$, there exists a finite $t > 0$ such that the knowledge of the input and the output over $[0, t]$ suffices to determine uniquely the initial state $\mathbf{x}(0)$. Otherwise, the system is said to be (globally) unobservable.

Note that this is a concept of deterministic observability taking no account of noises. Whatever estimation techniques are to be used, observability analysis is necessary that tells the inherent estimability of the system state [15, 16].

*Theorem*: If the matrix $\int_0^t \mathbf{M}^T \mathbf{M} dt$ is nonsingular and $\mathbf{C}_i^{m(0)}$ is restricted to be an identical matrix, then the system state is globally observable. (See below for the definition of the matrix $\mathbf{M}$)

Proof. Taking time derivative of the observation in (9)

$$\dot{\hat{\mathbf{m}}}^m = \dot{\mathbf{C}}_i^m \mathbf{m}^i = \hat{\mathbf{m}}^m \times (\mathbf{K} \mathbf{y}_g + \boldsymbol{\varepsilon})$$
$$= (\mathbf{y}_g^T \otimes \hat{\mathbf{m}}^m \times) vec(\mathbf{K}) + \hat{\mathbf{m}}^m \times \boldsymbol{\varepsilon}$$
$$= \begin{bmatrix} \mathbf{y}_g^T \otimes \hat{\mathbf{m}}^m \times & \hat{\mathbf{m}}^m \times \end{bmatrix} \begin{bmatrix} vec(\mathbf{K}) \\ \boldsymbol{\varepsilon} \end{bmatrix} \quad (10)$$
$$\triangleq \mathbf{M} \begin{bmatrix} vec(\mathbf{K}) \\ \boldsymbol{\varepsilon} \end{bmatrix}$$

where the operator $\otimes$ denotes the Kronecker product and $vec(\mathbf{K})$ forms a vector by stacking the columns of the matrix $\mathbf{K}$. The matrix equality $vec(\mathbf{ABC}) = (\mathbf{C}^T \otimes \mathbf{A}) vec(\mathbf{B})$ has been used above. Left multiplying $\mathbf{M}^T$ and integrating from zero time to current time, the gyroscope parameters can be solved by

$$\begin{bmatrix} vec(\mathbf{K}) \\ \boldsymbol{\varepsilon} \end{bmatrix} = \left( \int_0^t \mathbf{M}^T \mathbf{M} dt \right)^{-1} \int_0^t \mathbf{M}^T \dot{\hat{\mathbf{m}}}^m dt \quad (11)$$

when the matrix $\int_0^t \mathbf{M}^T \mathbf{M} dt$ is nonsingular.

However, the initial values for $\mathbf{C}_i^m$ and $\mathbf{m}^i$ cannot be uniquely determined, as for any attitude matrix $\mathbf{Q}$ it is always valid for the magnetometer observation

$$\hat{\mathbf{m}}^m = (\mathbf{C}_i^m \mathbf{Q})(\mathbf{Q}^T \mathbf{m}^i) \quad (12)$$

which means that the initial values of $\mathbf{C}_i^m$ and $\mathbf{m}^i$ both have infinite feasible solutions. Therefore, we restrict $\mathbf{C}_i^{m(t)}\big|_{t=0}$ to be an identity matrix, i.e., designating the initial magnetometer frame as the inertial frame, so as to make the formulation fully observable. Then the attitude $\mathbf{C}_i^m$ will be available with the determined gyroscope parameters by integrating (8). The constant magnetic vector is computed as $\mathbf{m}^i = \mathbf{C}_m^i \hat{\mathbf{m}}^m$ at any time.

∎

IV. SIMULATION AND TESTS

The error-state extended Kalman filter (EKF) is employed to carry out the state estimation for the state-space model (9). Deriving the corresponding first-order error-state equation is straightforward and we omitted here for brevity [6, 10]. A good initial state estimate is important to get satisfying EKF performance. The magnetometer orientation and local magnetic vector are respectively initialized by the identity matrix and the first magnetometer measurement $\hat{\mathbf{m}}^m(0)$. The initial estimate of $\mathbf{K}$ and $\boldsymbol{\varepsilon}$ could be obtained from (10) using the similar technique with [1]. Note that (11) is not directly usable as it needs to compute the magnetometer measurement derivative. Specifically, integrating (10) over the time interval $[t_k \ t_{k+1}]$,

$$\hat{\mathbf{m}}^m(t_{k+1}) - \hat{\mathbf{m}}^m(t_k)$$
$$= \int_{t_k}^{t_{k+1}} \mathbf{M} dt \cdot \begin{bmatrix} vec(\mathbf{K}) \\ \boldsymbol{\varepsilon} \end{bmatrix}, \quad k = 0, 1, 2 \ldots \quad (13)$$

which can be used to get a least-square estimate. Readers can refer to [1] for further details.

*A. Simulation Results*

We designed a simulator to generate the magnetometer and gyroscope measurements under attitude motion for 100 seconds. Figure 2 plots the orientation trajectory of the attitude motion in Euler angles (degree) and the generated true sensor measurements. The magnetometer is assumed having been well calibrated so as to examine the full potential of the proposed approach in calibrating gyroscope by the aid

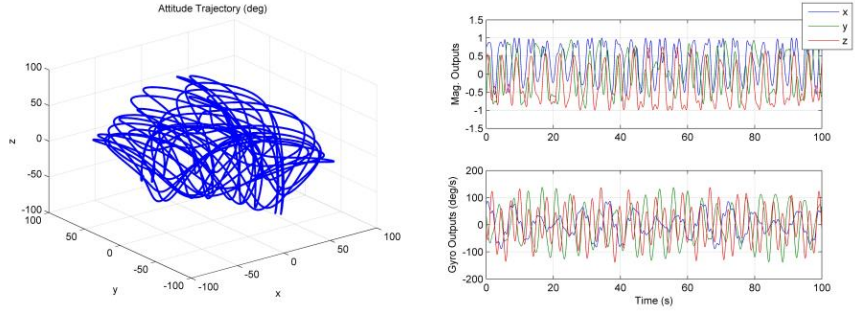

Figure 2. Orientation trajectory (left) and sensor measurements (right) in simulations.

Table I. Simulation Parameters and Calibration Results Across 50 Monte Carlo Runs

| | Simulation Parameters | True Values | Calibration Results | |
|---|---|---|---|---|
| | | | mean | std |
| Gyroscope | $\mathbf{K}_g$ | $\begin{bmatrix} 1.1 & 0.1 & 0.15 \\ & 1.2 & 0.2 \\ & & 1.3 \end{bmatrix}$ | $\begin{bmatrix} 1.1000 & 0.0997 & 0.1488 \\ & 1.2001 & 0.2000 \\ & & 1.3000 \end{bmatrix}$ | $\begin{bmatrix} 3.0 & 2.6 & 5.5 \\ & 1.4 & 3.0 \\ & & 2.6 \end{bmatrix} \times 10^{-4}$ |
| Gyroscope | $\boldsymbol{\varepsilon}_b$ (deg/s) | $\begin{bmatrix} 1 & 3 & 2 \end{bmatrix}^T$ | $\begin{bmatrix} 0.997 & 3.002 & 1.997 \end{bmatrix}$ | $\begin{bmatrix} 5.8 & 3.6 & 4.5 \end{bmatrix} \times 10^{-3}$ |
| Mag-Gyro Misalignment | $dcm2eul(\mathbf{C}_b^m)$ (deg) | $\begin{bmatrix} 10 & 20 & 15 \end{bmatrix}^T$ | $\begin{bmatrix} 9.993 & 19.969 & 15.031 \end{bmatrix}$ | $\begin{bmatrix} 0.008 & 0.018 & 0.022 \end{bmatrix}$ |

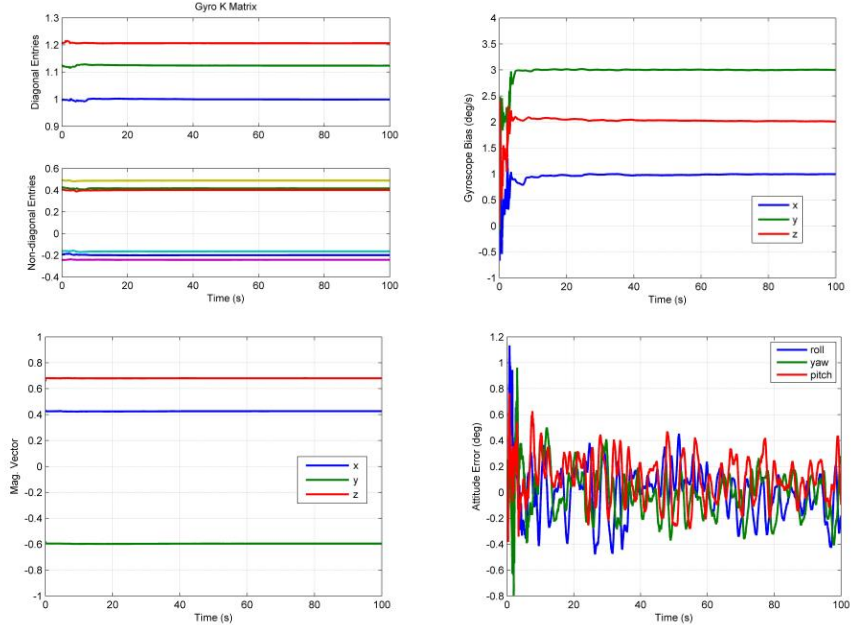

Figure 3. Calibration parameter estimates in a typical run. (Upper-left: matrix $\mathbf{K}$, upper-right: gyroscope bias $\boldsymbol{\varepsilon}_b$, lower-left: local magnetic vector $\mathbf{m}^i$, lower-right: angle error of $\mathbf{C}_i^m$)

of magnetometer. All simulated noises are subject to normal distribution, the standard deviation of magnetometer measurement noise is set to 0.01 (unitless) and the gyroscope noise density of is set to $0.02 \text{ deg/s}/\sqrt{Hz}$.

Table I lists the simulation parameters, and the mean and standard deviation of the calibration results across 50 Monte Carlo runs. The calibration parameters are determined to a high accuracy: gyroscope scale factor (900 ppm, 3$\sigma$), gyroscope non-orthogonality (0.09 deg, 3$\sigma$), gyroscope bias (0.02 deg/s, 3$\sigma$) and magnetometer-gyroscope misalignment (0.06 deg, 3$\sigma$). Figure 3 gives the transient behavior of the parameter estimates in a typical run. The parameters have very good convergence property. Though lacking of an absolute attitude reference, the attitude errors stay below 0.5 deg throughout the calibration procedure (lower-right subfigure). The fluctuation is caused by inevitable noises that are not considered by the observability analysis.

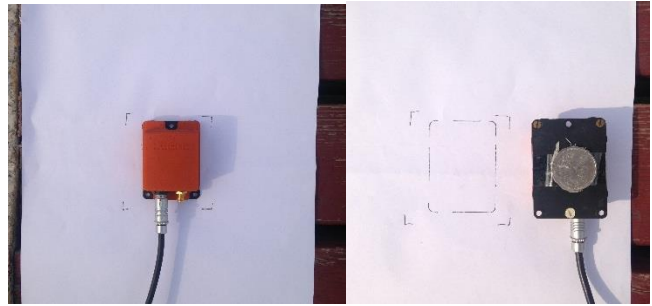

Figure 4. Xsens MTi-G-700 and RMB coin attached.

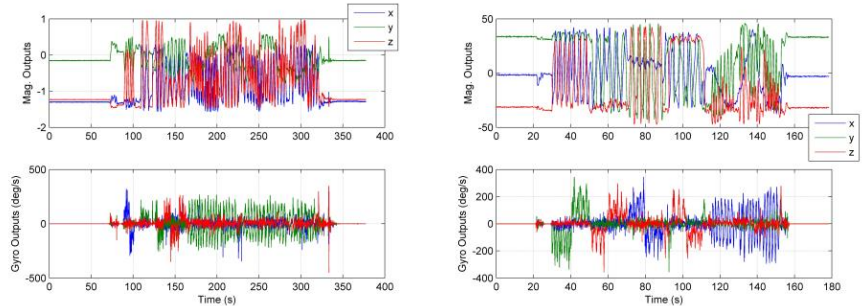

Figure 5. Sample outputs of Xsens MTi-G-700 (left, dataset #3) and iPhone-6 (right, dataset #3).

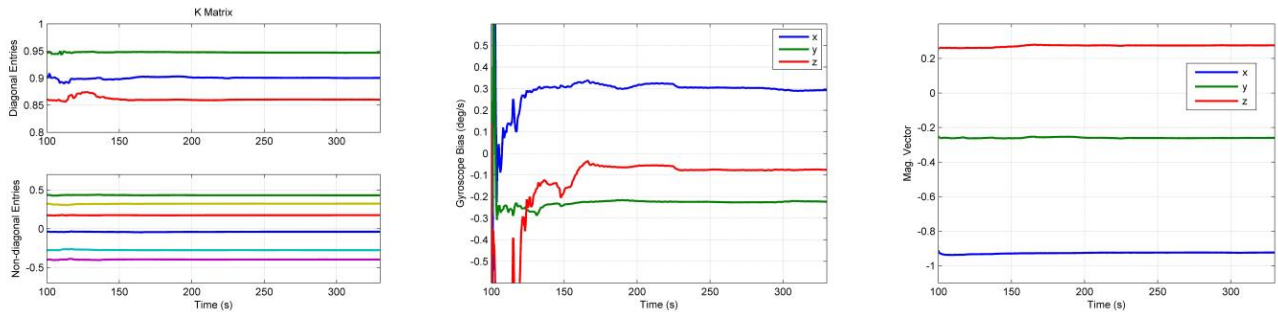

Figure 6. Calibration parameter estimates for MTi-G-700 dataset #3.
(Left: matrix **K**, middle: gyroscope bias $\varepsilon_b$, right: local magnetic vector $\mathbf{m}^i$)

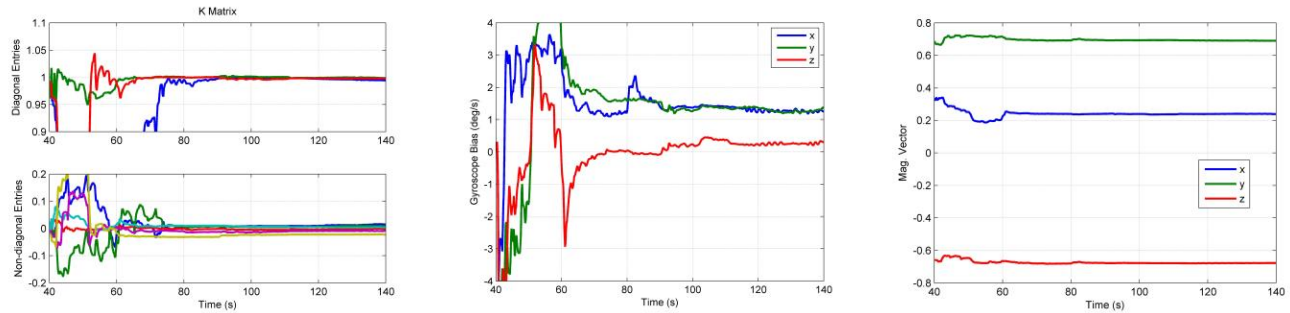

Figure 7. Calibration parameter estimates for iPhone-6 dataset #3.
(Left: matrix **K**, middle: gyroscope bias $\varepsilon_b$, right: local magnetic vector $\mathbf{m}^i$)

### B. Test Results

We use four datasets collected respectively from Xsens MTi-G-700 and iPhone-6 to evaluate the proposed calibration method. The MTi-G-700 datasets come from [1]. Two datasets (#1 and #2) were collected using the

Table II. Calibration Test Results

| Test Unit | Datasets | $\mathbf{K}_g$ | $\boldsymbol{\varepsilon}_b$ (deg/s) | $dcm2eul(\mathbf{C}_b^m)$ (deg) |
|---|---|---|---|---|
| **Xsens Mti-G-700** | #1 | $\begin{bmatrix} 0.9990 & 0.0004 & 0.0001 \\ & 1.0000 & 0.0006 \\ & & 0.9998 \end{bmatrix}$ | $[-0.232 \ \ 0.172 \ \ 0.248]^T$ | $[-0.026 \ \ 0.014 \ \ -0.111]^T$ |
| | #2 | $\begin{bmatrix} 0.9996 & 0.0006 & 0.0005 \\ & 0.9998 & -0.0007 \\ & & 1.0004 \end{bmatrix}$ | $[-0.215 \ \ 0.156 \ \ 0.235]^T$ | $[-0.017 \ \ -0.025 \ \ -0.204]^T$ |
| | #3 | $\begin{bmatrix} 0.9989 & \mathbf{0.0019} & 0.0008 \\ & 1.0017 & -0.0011 \\ & & \mathbf{1.0011} \end{bmatrix}$ | $[-0.241 \ \ 0.140 \ \ \mathbf{0.253}]^T$ | $[16.32 \ \ 23.78 \ \ \mathbf{9.97}]^T$ |
| | #4 | $\begin{bmatrix} 0.9975 & \mathbf{-0.0017} & 0.0001 \\ & 0.9985 & -0.0016 \\ & & \mathbf{0.9968} \end{bmatrix}$ | $[-0.211 \ \ 0.168 \ \ \mathbf{0.218}]^T$ | $[16.31 \ \ 23.88 \ \ \mathbf{10.16}]^T$ |
| | *Discrepancy* | *4000ppm (scale factor) 0.2 deg (non-orthogonality)* | *0.03* | *0.2* |
| **iPhone-6** | #1 | $\begin{bmatrix} 0.9984 & \mathbf{0.0093} & 0.0080 \\ & 1.0011 & 0.0019 \\ & & 0.9976 \end{bmatrix}$ | $[-1.16 \ \ -1.39 \ \ -0.20]^T$ | $[-1.49 \ \ 0.87 \ \ -1.22]^T$ |
| | #2 | $\begin{bmatrix} \mathbf{0.9931} & -0.0016 & 0.0015 \\ & 1.0007 & 0.0029 \\ & & 1.0010 \end{bmatrix}$ | $[\mathbf{-1.62} \ \ -1.37 \ \ -0.06]^T$ | $[-1.26 \ \ \mathbf{0.54} \ \ -0.89]^T$ |
| | #3 | $\begin{bmatrix} 0.9997 & \mathbf{-0.0019} & 0.0015 \\ & 0.9994 & -0.0016 \\ & & 1.0007 \end{bmatrix}$ | $[\mathbf{-1.10} \ \ -1.19 \ \ -0.14]^T$ | $[-1.26 \ \ 0.68 \ \ -0.75]^T$ |
| | #4 | $\begin{bmatrix} \mathbf{1.0012} & 0.0010 & 0.0104 \\ & 0.9986 & 0.0004 \\ & & 1.0024 \end{bmatrix}$ | $[-1.11 \ \ -1.18 \ \ 0.10]^T$ | $[-1.41 \ \ \mathbf{1.06} \ \ -1.16]^T$ |
| | *Discrepancy* | *8000ppm (scale factor) 0.6 deg (non-orthogonality)* | *0.5* | *0.5* |

Note: the peaks are highlighted in bold.

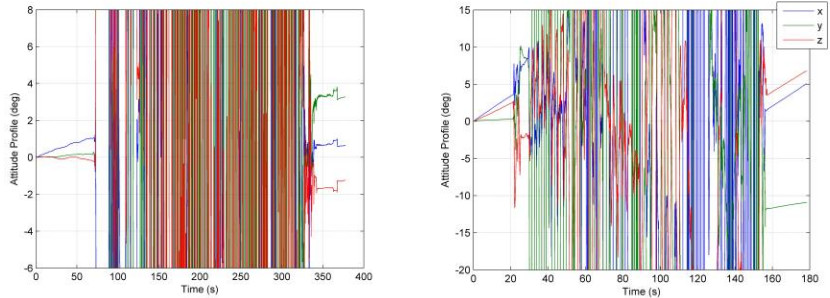

Figure 8. Dead-reckoning attitude result using calibrated gyroscope measurement of MTi-G-700 dataset #3 (left) and iPhone-6 dataset #3 (right).

MTi-G-700 unit alone (Fig. 4, left), while the other two datasets (#3 and #4) were collected with a RMB coin taped onto the unit bottom plate. The coin is made of soft-iron magnetic material (Fig. 4, right). The iPhone-6 datasets were collected similarly but with no coin attached. Typical raw sensor measurements of MTi-G-700 and iPhone-6 are plotted in Fig. 5. In all datasets, the test units were kept stationary on the ground with exactly the same pose at the start and end of the tests.

Their calibration results are both listed in Table II.

For the MTi-G-700 unit, as the coin affects the magnetometer frame, the magnetometer-gyroscope misalignment attitude is significantly changed [1]. As expected, the gyroscope parameters is apparently immune to the coin attachment. For MTi-G-700, the peak-to-peak discrepancy among four sets of calibration parameters is up to 4000 ppm for the gyroscope scale factor, 0.2 deg for the gyroscope non-orthogonality, 0.03 deg/s for the gyroscope bias and 0.2 deg for the magnetometer-gyroscope misalignment, while for iPhone-6 it is up to 8000 ppm for the gyroscope scale factor, 0.6 deg for the gyroscope non-orthogonality, 0.5 deg/s for the gyroscope bias and 0.5 deg for the magnetometer-gyroscope misalignment. Figure 6 plots the estimates of the calibration parameters for the MTi-G-700 dataset #3, while Fig. 7 plots the estimates of the calibration parameters for the iPhone-6 dataset #3. Note the stationary data at the start and end of all datasets has been excluded from the calibration data. We observed that the calibration estimated does not converge sufficiently until all three axes have experienced significant rotation excitation. For example, in the iPhone-6 dataset #3 (Fig. 5, right), the third axis performs significant rotation at about 70s, which accords with convergence time in Fig. 7.

As the true calibration parameters are unknown, we use the fact that the tests started and ended at the same pose to indirectly evaluate the calibration quality. Figure 8 presents the dead-reckon attitude result using the calibrated gyroscope measurements of the MTi-G-700 dataset #3 and the iPhone-6 dataset #3. The MTi-G-700 unit drifts about 3.5 deg in 360 seconds and the iPhone-6 unit drifts about 12 deg in 180 seconds.

## V. CONCLUSIONS

A reference input is a fundamental requirement of any sensor calibration. Existing gyroscope calibration methods mostly rely on external apparatus to provide reference inputs. This paper proposes an on-site gyroscope calibration method with the help of magnetometer in the homogeneous magnetic field. It is shown that, with sufficient rotation excitation, the homogeneous magnetic field vector can be exploited to serve as a good reference for calibrating low-cost gyroscopes. The gyroscope calibration parameters include the scale factor, non-orthogonal coefficient and bias of three axes, and its misalignment to the magnetometer frame. Simulation and field test results using Xsens MTi and iPhone 6 demonstrate the effectiveness of the method.